\documentclass[11pt,a4paper]{article}
\usepackage[hyperref]{emnlp-ijcnlp-2019}
\usepackage{times}
\usepackage{latexsym}

\usepackage{url}

\usepackage{amsmath}
\usepackage{amsfonts}
\usepackage{xcolor}
\usepackage{graphicx}
\usepackage{verbatim}
\usepackage{multirow}

\aclfinalcopy 

\newcommand{\abc}[1]{\textcolor{black}{#1}}
\newcommand{\wqz}[1]{\textcolor{black}{#1}}

\title{Towards Diverse and Accurate Image Captions via Reinforcing Determinantal Point Process}

\author{Qingzhong Wang \and Antoni B.Chan\\
  Department of Computer Science, City University of Hong Kong \\
  {\tt qingzwang2-c@my.cityu.edu.hk, abchan@cityu.edu.hk} 
  }

\date{}

\begin{document}
\maketitle
\begin{abstract}
Although significant progress has been made in the field of automatic image captioning, it is still a challenging task. Previous works normally pay much attention to improving the quality of the generated captions but ignore the diversity of captions. 
In this paper, we combine determinantal point process (DPP) and reinforcement learning (RL) and propose a novel reinforcing DPP (R-DPP) approach to generate a set of captions with high quality and diversity for an image. We show that R-DPP performs better on accuracy and diversity than using noise as a control signal (GANs, VAEs). Moreover, R-DPP is able to preserve the modes of the learned distribution. Hence, beam search algorithm can be applied to generate a single accurate caption, which performs  better than other RL-based models.
\end{abstract}

\section{Introduction}\label{sec1:intro}

Image captioning, which combines the fields of computer vision (CV) and natural language processing (NLP), is a challenging task, which has drawn much attention from the two communities
and significant progress has been achieved. Earlier works \cite{babytalk,templatemodel,visconcept} generally directly employ vision and language models. 
However, these two-stage models cannot be trained in a end-to-end manner, which limits their performance.

Recently, CNN-LSTM models have become  popular \cite{NIC, spatt}.
CNN-LSTM models are typically composed of three modules: (1) a visual CNN, (2) a \abc{language} LSTM, and (3) the connection module between them, which can be trained in an end-to-end manner. 
More powerful captioning models are later proposed \cite{bottomup,SCST,hieratt}, and trained using reinforcement learning (RL) \abc{where the evaluation metric (e.g., CIDEr) is used as the reward function.}  As a result, the generated captions obtain  high quality according to the most popular metrics, such as BLEU \cite{bleu}, METEOR \cite{M}, ROUGLE \cite{R}, CIDEr \cite{C} and SPICE \cite{spice}.

However, most of the above models do not focus on the diversity of captions.  While directly maximizing the metrics using RL \cite{SCST} significantly improves the metric scores, they lack diversity even though they are randomly drawn from the learned distribution  \cite{my-div-paper}. 
The lack of diversity in the captions is further exacerbated when using beam search to find the mode of the learned distribution. 

 The main issue of RL-based methods that leads to generating less diverse captions is they only consider the quality (as measured by BLEU or CIDEr)
 of samples during training.
To address this issue, in this paper, we propose a novel approach that combines RL and determinantal point processes (DPP) \cite{DPP-uai} that generates both accurate and diverse image captions.
Inspired by DPPs, which account for the quality and diversity of subsets, we first propose a new metric that is able to reflect the quality and diversity of a set of captions.  We then maximize the proposed  metric score using RL, which is equivalent to a  DPP training process.
We evaluate our model using the  diversity metrics from \cite{my-div-paper}, and our proposed R-DPP model achieves both high accuracy and high diversity scores. In addition, R-DPP preserves the modes of the learned distribution -- applying the beam search algorithm to generate one high-quality caption yields better performance than the baseline captioning model.
Moreover, R-DPP outperforms its counterparts on the oracle test (see Table \ref{table2}).

\section{Related Work}\label{sec2:related}

\paragraph{Diverse image captioning.} Recently, generating diverse captions receives much attention, and a variety of captioning models are developed, such as CVAE \cite{cvae}, CGAN \cite{cgan}, GroupTalk \cite{grouptalk}, GroupCap \cite{groupcap}, POS \cite{posg} and SCT \cite{SCT}. CVAE and CGAN employ random noise vectors to control the difference among the generated captions. However, the diversity is highly related to the variance of the noise, which makes it difficult to balance diversity and accuracy. GroupTalk employ multiple captioners\footnote{A captioner could be a captioning model.} and a classifier to generated diverse captions. Each captioner generate one caption and the classifier is used to control the diversity among the captions. However the computational cost is high due to its use of multiple captioners.
 GroupCap considers the structure relevance and diversity constraint to generate both accurate and diverse captions, in which VP-trees are constructed. POS introduces part-of-speech (POS) tags to control the difference among captions, which contains two branches: 1) POS tag prediction, 2) word prediction. The same POS tag could result in using different words (synonyms), leading to diversity. Instead of employing POS tags as control signals, SCT applies noun chunks that are obtained by dependency parsing \cite{chen2014fast}.
Compared with the above captioning models, our proposed RL using DPP is much simpler and more efficient,  does not require any other branches or control signals, \abc{and can be applied to  any baseline captioning model.}

\paragraph{Determinantal point process (DPP).} Given a discrete set $\mathcal{X} = \{x_1, x_2, \cdots, x_N\}$, a DPP $\mathcal{P}$ measures the probability of each subset $\mathbf{X}$ of $\mathcal{X}$, which is defined as \cite{DPP-uai}:
\begin{align}\label{eq1}
\mathcal{P}_L(\mathbf{X})=\tfrac{\textbf{det}(L_{\mathbf{X}})}{\textbf{det}(L+I)},
\end{align}
where $L$ is a positive semidefinite matrix, representing an \textit{L-ensemble}, $I$ denotes the $N\times N$ identity matrix and $\textbf{det}(L+I)=\sum_{\mathbf{X}\subseteq \mathcal{X}} \textbf{det}(L_{\mathbf{X}})$.

Generally, $L=[L_{ij}]$ can be decomposed as a Gram matrix with elements
$L_{ij}=q_i\phi_i^T\phi_jq_j$, 
where $q_i$ denotes the quality of the $i$th element and $s_{ij}=\phi_i^T\phi_j$ denotes the similarity between the $i$th and $j$th elements, where $||\phi_i||=1$.

A DPP is trained by maximizing the log-likelihood $\log \mathcal{P}_L(\mathbf{X})$, where the subset with larger $\textbf{det}(L_{\mathbf{X}})$ will be assigned a higher probability. \abc{Inference involves finding the subset with highest posterior probability (MAP).}
DPP has been used in applications that require both quality and diversity: 
text summarization \cite{DPP-uai}, video summarization \cite{video-dpp}, recommendation \cite{fast-dpp} and neural conversation \cite{conversation-dpp}.

\section{DPP-based Reinforcement Learning for Image Captioning}\label{sec3:method}

We consider each caption as an item, and define the quality  of a caption using CIDEr,
\begin{align}\label{eq3}
q_i=\textbf{CIDEr}(c_i, \mathcal{C}_{GT}),
\end{align}
where $c_i$ denotes the $i$th caption in a subset, $\mathcal{C}_{GT}$ denotes human annotations and $\textbf{CIDEr}(\cdot, \cdot)$ is the CIDEr score. We define the similarity 
between captions as (i.e., ``self-CIDEr'' in  \cite{my-div-paper}), 
\begin{align}\label{eq4}
s_{ij} = \textbf{CIDEr}(c_i, c_j).
\end{align}
The $L$ matrix in DPP is then 
\begin{align}\label{eq5}
L = \mathbf{q}^T\mathbf{q}\odot\mathbf{S},
\end{align}
where $\mathbf{q}=[q_1, \cdots, q_N]$, 
$\mathbf{S}=[s_{ij}]$, 
 and $\odot$ denotes element-wise multiplication.

Let $M(\theta)$ be the captioning model and $\mathcal{C}=\{c_1, c_2, \cdots, c_m\}$ a subset of $m$ captions sampled from $M(\theta)$. The probability of $\mathcal{C}$ can be measured with (\ref{eq1}), using the determinants of $L_{\mathcal{C}}$ and $L+I$. Unfortunately, to compute $L$ is intractable since the number of possible captions $N$ is huge, roughly $|D|^{l_m}$, where $|D|$ is the dictionary size (10,000) and $l_m$ is the caption length (16).
%
%
Although 
computing $L$ is intractable, 
\abc{we note that $L$ is a constant w.r.t.~$\theta$ for a fixed dictionary $D$ and  caption length $l_m$.
Thus, the denominator in (\ref{eq1}) can be ignored when maximizing the likelihood of the generated captions $\cal C$ w.r.t.~$\theta$,}
\begin{align}
\nonumber
\theta^*=\mathop{\mathrm{argmax}} \mathcal{P}_L(\mathcal{C}) 
=\mathop{\mathrm{argmax}} \log(\textbf{det}(L_{\mathcal{C}})).
\end{align}
To compute the quality scores and similarity matrix, we should sample a set of captions $\mathcal{C}$ from $M(\theta)$, and thus we cannot directly calculate the gradient of $\log(\textbf{det}(L_{\mathcal{C}}))$ w.r.t. $\theta$. Alternatively, we can first compute the derivative 
$\tfrac{\partial\log(\textbf{det}(L_{\mathcal{C}}))}{\partial L^{\mathcal{C}}_{ij}} = \hat{L}^{\mathcal{C}}_{ij}$,
where $L^{\mathcal{C}}_{ij}$ is the element of $L_{\mathcal{C}}$ and $\hat{L}^{\mathcal{C}}_{ij}$ is the element of $L_{\mathcal{C}}^{-1}$.\footnote{Adding a  small constant $\epsilon I$ to  $L_{\mathcal{C}}$  ensures invertability.}
Considering the derivative $\hat{L}^{\mathcal{C}}_{ij}$, its sign indicates whether we should reduce or increase $L^{\mathcal{C}}_{ij}$ to enlarge $\log(\textbf{det}(L_{\mathcal{C}}))$.

 Recall that the reward function in \cite{SCST} is the expectation of CIDEr, 
\begin{align}\label{eq8}
R(\theta)=\sum_{i=1}^m q_i p_{\theta}(c_i),
\end{align}
and the corresponding policy gradient is 
\begin{align}\label{eq9}
\nabla_{\theta}R(\theta)=\sum_{i=1}^m q_i \nabla_{\theta}\log(p_{\theta}(c_i))\cdot p_{\theta}(c_i).
\end{align}
(\ref{eq9}) shows that the probability of the high-quality captions will increase, and finally the model could tend to generate captions that have high quality but lack diversity.

The main issue of using (\ref{eq8}) is that it only accounts for the quality of captions. To promote diversity, we employ a new reward function that considers each pair of captions in $\mathcal{C}$, 
\begin{align}\label{e10}
R(\theta)=\sum_{i=1}^m\sum_{j=1}^m \textbf{sign}(\hat{L}^{\mathcal{C}}_{ij}) L^{\mathcal{C}}_{ij} p_{\theta}(c_i)p_{\theta}(c_j),
\end{align}
where $\textbf{sign}(x)$ is the sign of $x$, 
and $p_{\theta}(c_i)$ is the probability of the $i$th caption according to $M(\theta)$. Note that $p_{\theta}(c_i)p_{\theta}(c_j)$ is the joint probability of the $i$th and $j$th captions, since the captions are sampled independently. Our reward function considers both the quality of captions as well as the similarity among captions (see Eq. (\ref{eq5}))\footnote{Note that the expectation of $L_{ij}^{\mathcal{C}}$ could be enlarged or reduced based on $\textbf{sign}(\hat{L}^{\mathcal{C}}_{ij})$, which is different with Eq. (\ref{eq8}) where the expectation of $q_i$ is always enlarged.}, thus is able to 
balance the quality and diversity.
The corresponding policy gradient is 
\wqz{(see supplemental for derivation)}:
%
{\small
\begin{equation}\label{eq11}
\begin{aligned}
& \nabla_{\theta}R(\theta)= \\
& 2\sum_{i=1}^m\nabla_{\theta}\log(p_{\theta}(c_i))p_{\theta}(c_i)\underbrace{\sum_{j=1}^m\textbf{sign}(\hat{L}^{\mathcal{C}}_{ij})L^{\mathcal{C}}_{ij}p_{\theta}(c_j)}_{\mathbb{E}\left[\textbf{sign}(\hat{L}^{\mathcal{C}}_{ij})L^{\mathcal{C}}_{ij}\right]},
\end{aligned}
\end{equation}
}which has the same form as (\ref{eq9}), but here we consider both quality and similarity among captions.


\section{Experiments}\label{sec4:experiments}

\paragraph{Experimental setup.}~We conduct our experiments on MSCOCO dataset, 
which has 123,287 annotated images, each with at least 5 captions.  Following  \cite{SCST}, we use 5k images for validation, 5k for testing and the remaining for training.
%
Our baseline captioning model is based on Att2in \cite{SCST}. We first train the model for 100 epochs using cross-entropy loss, and then refine it for another 100 epochs using our policy gradient in (\ref{eq11}). During training, we apply Adam with learning rate 0.0004.
For comparison, we also refine the baseline model for 100 epochs using original policy gradient in (\ref{eq9}). We also compare 
with CGAN\footnote{We train CGAN without using rollout, which is different from \cite{cgan}}, GMM-CVAE \cite{cvae}, SCST \cite{SCST}, and XE+$\lambda$CIDEr \cite{my-div-paper}.
The diversity metric is self-CIDEr diversity, which is shown to be more correlated to human judgment \cite{my-div-paper}.

\paragraph{Results.} 

\begin{figure}[t]
\centering
\includegraphics[width=0.5\textwidth]{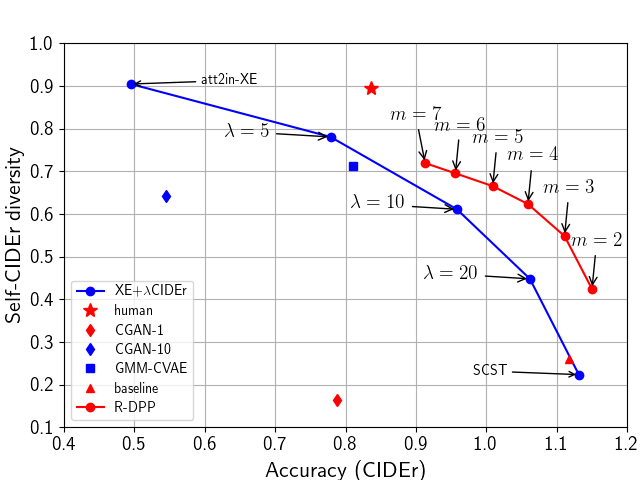}
\caption{Performance on diversity and accuracy. The captions are generated via random sampling from the learned distribution. For each model we sample 10 captions to compute the self-CIDEr diversity scores \cite{my-div-paper}, and the accuracy score is the average of CIDEr scores. CGAN-\{1,10\} use standard deviations of 1 and 10 to train CGANs, and greedy search is used for inference. $m$ is the number of samples used to train our R-DPP.}\label{fig1}
\end{figure}

\begin{table}[t]
\centering
\scalebox{0.6}{
\begin{tabular}{@{}c|c|ccccc@{}}
\hline
\textbf{Model} &$bw$ &B-4 &M &R &C &S \\
\hline
Adaptive-XE \cite{when2look} &3 &0.332 &0.266 &- &1.085 &- \\
Updown-XE \cite{bottomup} &5 &0.362 &0.270 &0.564 &1.135 &0.203 \\
Updown-RL \cite{bottomup} &5 &0.363 &0.277 &0.569 &1.201 &0.214 \\
DISC-RL \cite{disccap} &2 &0.363 &0.273 &0.571 &1.141 &0.211 \\
Hieratt-XE \cite{hieratt} &3 &0.362 &0.275 &0.566 &1.148 &0.206 \\
Hieratt-RL \cite{hieratt} &3 &0.376 &0.278 &0.581 &1.217 &0.215 \\
\hline
SCST \cite{SCST} &- &0.333 &0.263 &0.553 &1.114 &- \\
Att2in-XE \cite{SCST} &- &0.313 &0.260 &0.543 &1.013 &- \\
XE+5CIDEr \cite{my-div-paper} &3 &0.382 &0.277 &0.579 &1.172 &0.206 \\
XE+10CIDEr \cite{my-div-paper} &3 &0.378 &0.276 &0.580 &1.174 &0.207 \\
XE+20CIDEr \cite{my-div-paper} &3 &0.375 &0.276 &0.579 &1.173 &0.209 \\
\hline
Our R-DPP ($m=2$) &3 &0.371 &0.279 &0.579 &\textbf{1.222} &0.214 \\
Our R-DPP ($m=3$) &3 &0.369 &0.278 &0.577 &1.216 &0.214 \\
Our R-DPP ($m=4$) &3 &0.360 &0.280 &0.572 &1.198 &0.214 \\
Our R-DPP ($m=5$) &3 &0.357 &0.278 &0.568 &1.179 &0.212 \\
Our R-DPP ($m=6$) &3 &0.352 &0.276 &0.566 &1.146 &0.208 \\
Our R-DPP ($m=7$) &3 &0.347 &0.272 &0.562 &1.124 &0.206 \\
\hline
\end{tabular}
}
\caption{Performance on single caption generation. The caption is generated using beam search ($bw$ is the beam width).  $m$ is the number of samples used during training of our R-DPP. The ``-XE'' suffix indicates training using cross-entropy loss, and ``-RL'' means fine-tuned with RL. \{B, M, R, C, S\} are abbreviations for BLEU, METEOR, ROUGE, CIDEr, and SPICE.}\label{table1}
\end{table}

\begin{table}[t]
\centering
\scalebox{0.63}{
\begin{tabular}{@{}c|c|cccc@{}}
\hline
\textbf{Model} &\# &B-4 &M &R &C \\
\hline
AG-CVAE \cite{cvae} &20 &0.471 &0.309 &0.638 &1.308 \\
GMM-CVAE \cite{cvae} &20 &0.449 &0.299 &0.624 &1.251 \\
POS \cite{posg} &20 &0.449 &0.365 &0.678 &1.468 \\
SCT \cite{SCT} &20 &0.448 &0.366 &0.689 &1.565 \\
\hline
SCST \cite{SCST} &20 &0.332 &0.322 &0.630 &1.383 \\
Att2in-XE \cite{SCST} &20 &0.329 &0.326 &0.621 &1.216 \\
XE+5CIDEr \cite{my-div-paper} &20 &0.462 &0.372 &0.682 &1.512 \\
XE+10CIDEr \cite{my-div-paper} &20 &0.465 &0.372 &0.689 &1.568 \\
XE+20CIDEr \cite{my-div-paper} &20 &0.427 &0.359 &0.673 &1.525 \\
\hline
\multirow{2}{*}{Our R-DPP ($m=2$)} &10 &0.407 &0.349 &0.659 &1.495  \\
&20 &0.443 &0.365 &0.677 &1.563 \\
\hline
\multirow{2}{*}{Our R-DPP ($m=3$)} &10 &0.442 &0.367 &0.677 &1.567  \\
&20 &0.494 &0.388 &0.702 &1.656 \\
\hline
\multirow{2}{*}{Our R-DPP ($m=4$)} &10 &0.455 &0.374 &0.686 &\textcolor{blue}{1.585}  \\
&20 &0.518 &0.400 &0.713 &1.691 \\
\hline
\multirow{2}{*}{Our R-DPP ($m=5$)} &10 &\textcolor{blue}{0.463} &\textcolor{blue}{0.375} &\textcolor{blue}{0.688} &\textcolor{blue}{1.585} \\
&20 &0.527 &\textbf{0.405} &0.717 &\textbf{1.700} \\
\hline
\multirow{2}{*}{Our R-DPP ($m=6$)} &10 &0.458 &0.374 &0.686 &\textcolor{blue}{1.585} \\
&20 &0.528 &0.403 &\textbf{0.718} &1.690 \\
\hline
\multirow{2}{*}{Our R-DPP ($m=7$)} &10 &0.452 &0.373 &0.683 &1.545 \\
&20 &\textbf{0.529} &0.404 &\textbf{0.718} &1.684 \\
\hline
\end{tabular}
}
\caption{Oracle (upper bound) performance based on each metric. \# represents the number of samples during inference. For Att2in-XE, XE+$\lambda$CIDEr, SCST and our R-DPP models, we randomly sample captions from the trained model and the results of other models are from their papers. The blue numbers are the highest scores when sample 10 captions and the bold ones are the highest scores when sample 20 captions.}\label{table2}
\end{table}

Fig.~\ref{fig1} shows the performance of different models in the diversity-accuracy space. Human annotations achieve relatively high diversity and accuracy\footnote{The accuracy score of human annotations is the leave-one-out CIDEr score as in \cite{my-div-paper}.}, and there is still a large gap between the proposed models and human annotations. Our R-DPP model slightly improves the accuracy of SCST and the baseline model (Att2in), when $m=2$, but the diversity score roughly doubles (0.2 to 0.4). Our R-DPP achieves comparable diversity scores as XE+$\lambda$CIDEr, but the captions generated by XE+$\lambda$CIDEr have lower accuracy compared to R-DPP. By maximizing $\textbf{det}(L_{\mathcal{C}})$, our R-DPP can simultaneously improves the quality and suppresses the similarity among captions (improves diversity). Comparing GMM-CVAE and R-DPP, both methods can generate captions with similar diversity, while R-DPP ($m=6$) has higher accuracy (0.8 vs 0.95),
which indicates that R-DPP better approximates the modes of the ground-truth distribution. Finally, the  R-DPP curve shows that the number of samples $m$ used during training balances the diversity and accuracy of the model. A larger $m$ leads to a more diverse set of captions, although it also incurs higher training computational cost. 

Another advantage of R-DPP is that it can be used to generate a single high-quality caption for an image. Table \ref{table1} shows the comparison between R-DPP and the state-of-the-art models. Compared with SCST, R-DPP improves the CIDEr score from 1.114 to 1.222, and the other metric scores are also improved by around 5\% or larger. Comparing with Hieratt-RL (state-of-the-art), R-DPP obtains similar CIDEr score, however, the Hieratt model cannot generate diverse captions. 

Fig.~\ref{fig1} and Table \ref{table1} show the effectiveness of R-DPP on generating both diverse and accurate captions, whereas they do not consider the optimal selection of $m$.
Hence, we conduct experiments on oracle test (see Table \ref{table2})---the upper bound of each metric. R-DPP outperforms other methods, providing the highest-quality caption based on generating 20 captions. Even sampling 10 captions, R-DPP obtains higher scores. With the increase of $m$, the scores increase in the beginning, but then fall, e.g., when we sample 20 captions, CIDEr score rises from 1.563 to 1.700 when $m$ increases from 2 to 5, after that it falls to 1.684. Also, when we sample 10 captions, R-DPP($m=5$) performs better. Thus, using $m=5$ could be a better choice to well balance diversity and accuracy, which also obtains the highest-quality caption.
We show more qualitative examples in the supplemental.

\section{Conclusion}

We have presented the reinforcing DPP (R-DPP) model, which is a simpler but efficient method for training a caption model to generate both diverse and accurate captions. Compared with other models, R-DPP obtains similar diversity score, but much higher accuracy score. In addition, the state-of-the-art oracle performance is significantly improved by R-DPP. In the future, we believe that more quality and diversity measurements should be introduced into R-DPP. It is also possible to extend R-DPP to other text generation tasks, such as dialog and machine translation, in order to provide diverse high-quality choices to the users.


\bibliography{myref}
\bibliographystyle{acl_natbib}

\newpage
\appendix
\onecolumn
The supplemental is arranged as follows:
\begin{itemize}
\item Details of the gradient computation.
\item Qualitative examples of diverse image captions.
\end{itemize}

\section{Gradient Computation}
We show how to compute the policy gradient in Eq. (11) in our paper. Recall that the reward function is defined as follows:
\begin{equation}\label{supp:eq1}
R(\theta)=\sum_{i=1}^m\sum_{j=1}^m \textbf{sign}(\hat{L}^{\mathcal{C}}_{ij}) L^{\mathcal{C}}_{ij} p_{\theta}(c_i)p_{\theta}(c_j).
\end{equation}
Note that only $p_{\theta}(\cdot)$ is a function of $\theta$, then we have
{\small
\begin{align}
\nabla_{\theta}R(\theta) = &\sum_{i=1}^m\sum_{j=1}^m\textbf{sign}(\hat{L}^{\mathcal{C}}_{ij})L^{\mathcal{C}}_{ij}\left(\nabla_{\theta}p_{\theta}(c_i)p_{\theta}(c_j) + \nabla_{\theta}p_{\theta}(c_j)p_{\theta}(c_i) \right) \\ 
= & \sum_{i=1}^m\sum_{j=1}^m\textbf{sign}(\hat{L}^{\mathcal{C}}_{ij})L^{\mathcal{C}}_{ij}\nabla_{\theta}p_{\theta}(c_i)p_{\theta}(c_j) + \sum_{i=1}^m\sum_{j=1}^m\textbf{sign}(\hat{L}^{\mathcal{C}}_{ij})L^{\mathcal{C}}_{ij}\nabla_{\theta}p_{\theta}(c_j)p_{\theta}(c_i) \\
= & \sum_{i=1}^m\sum_{j=1}^m\textbf{sign}(\hat{L}^{\mathcal{C}}_{ij})L^{\mathcal{C}}_{ij}\nabla_{\theta}\log(p_{\theta}(c_i))p_{\theta}(c_i)p_{\theta}(c_j) + \sum_{i=1}^m\sum_{j=1}^m\textbf{sign}(\hat{L}^{\mathcal{C}}_{ij})L^{\mathcal{C}}_{ij}\nabla_{\theta}\log(p_{\theta}(c_j))p_{\theta}(c_j)p_{\theta}(c_i) \\
=& 2\sum_{i=1}^m\sum_{j=1}^m\textbf{sign}(\hat{L}^{\mathcal{C}}_{ij})L^{\mathcal{C}}_{ij}\nabla_{\theta}\log(p_{\theta}(c_i))p_{\theta}(c_i)p_{\theta}(c_j) \\
=& 2\sum_{i=1}^m\nabla_{\theta}\log(p_{\theta}(c_i))p_{\theta}(c_i)\underbrace{\sum_{j=1}^m\textbf{sign}(\hat{L}^{\mathcal{C}}_{ij})L^{\mathcal{C}}_{ij}p_{\theta}(c_j)}_{\mathbb{E}\left[\textbf{sign}(\hat{L}^{\mathcal{C}}_{ij})L^{\mathcal{C}}_{ij}\right]}.
\end{align}\label{supp:eq2}
}
Since $L^{\mathcal{C}}$ and $\hat{L}^{\mathcal{C}}$ are symmetric mtrices, we can derive Eq.~(13) from Eq.~(12). Note that $\frac{\partial \log p_\theta}{\partial \theta} \equiv \frac{1}{p_\theta}\frac{\partial p_\theta}{\partial \theta}$ for $p_\theta>0$, hence, we obtain Eq~(12) from Eq.~(11).

\section{Qualitative Examples}
\begin{figure*}[t]
\centering
\includegraphics[width=\textwidth]{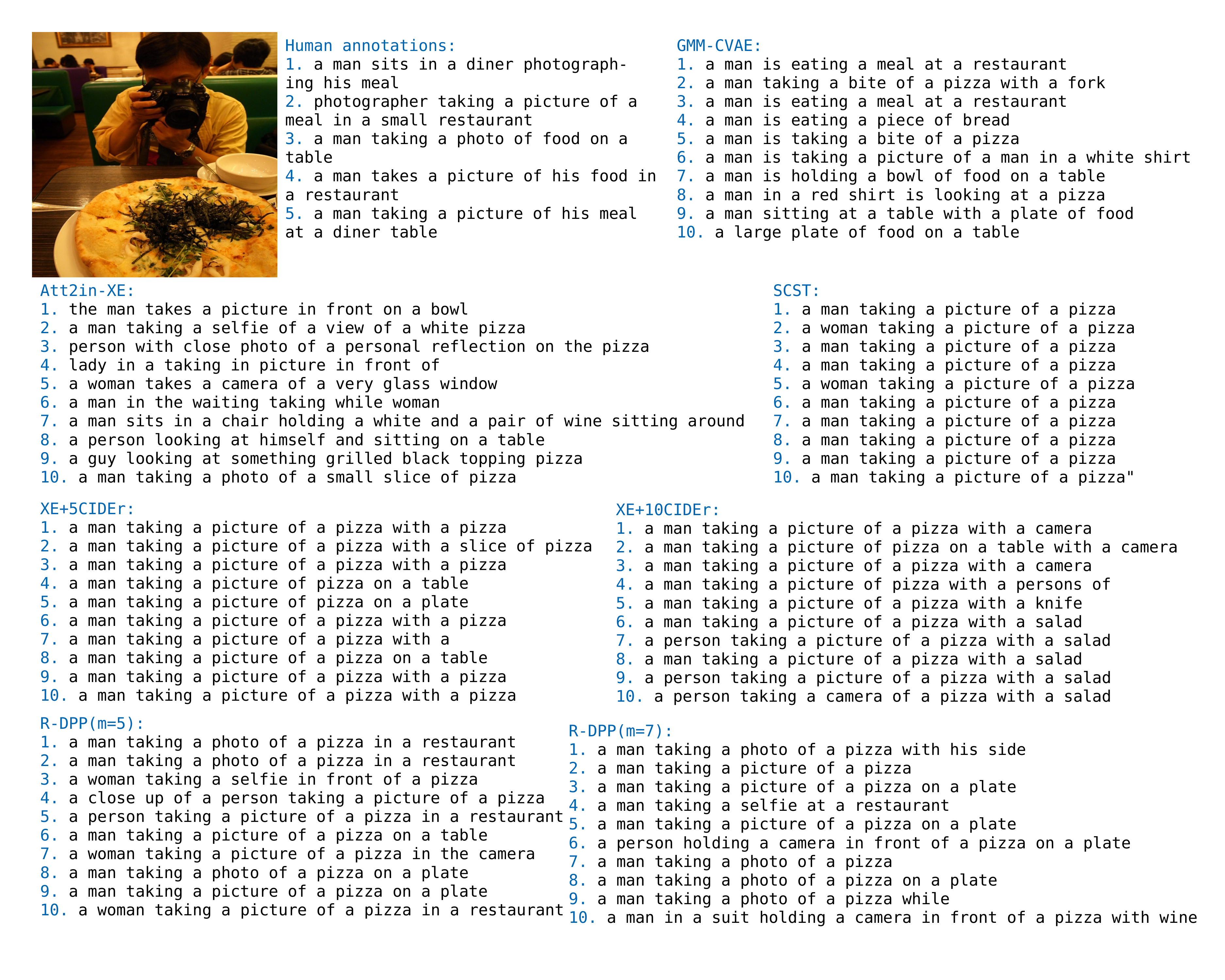}
\caption{•}\label{supp:fig1}
\end{figure*}

\begin{figure*}[t]
\centering
\includegraphics[width=\textwidth]{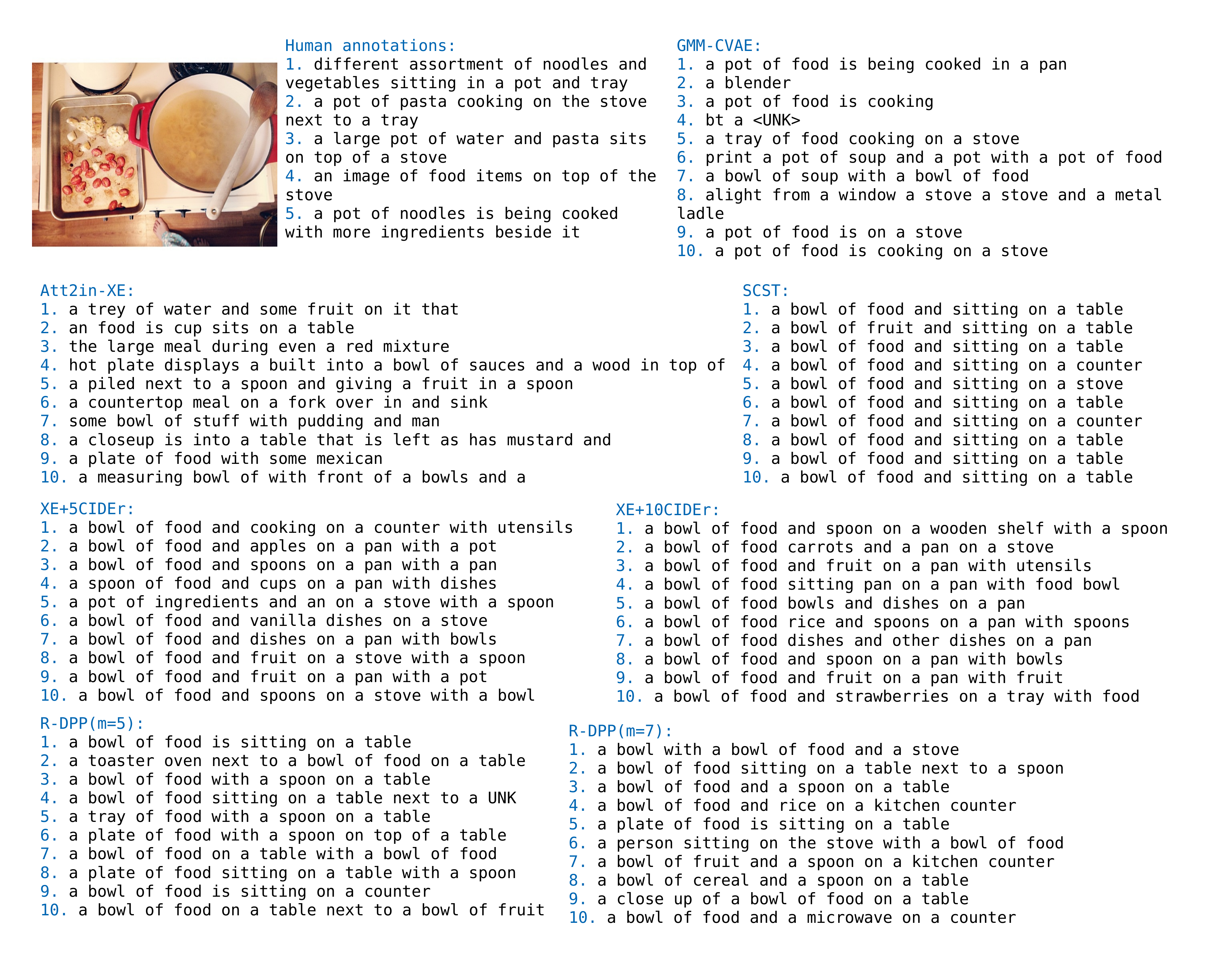}
\caption{•}\label{supp:fig2}
\end{figure*}

\begin{figure*}[t]
\centering
\includegraphics[width=\textwidth]{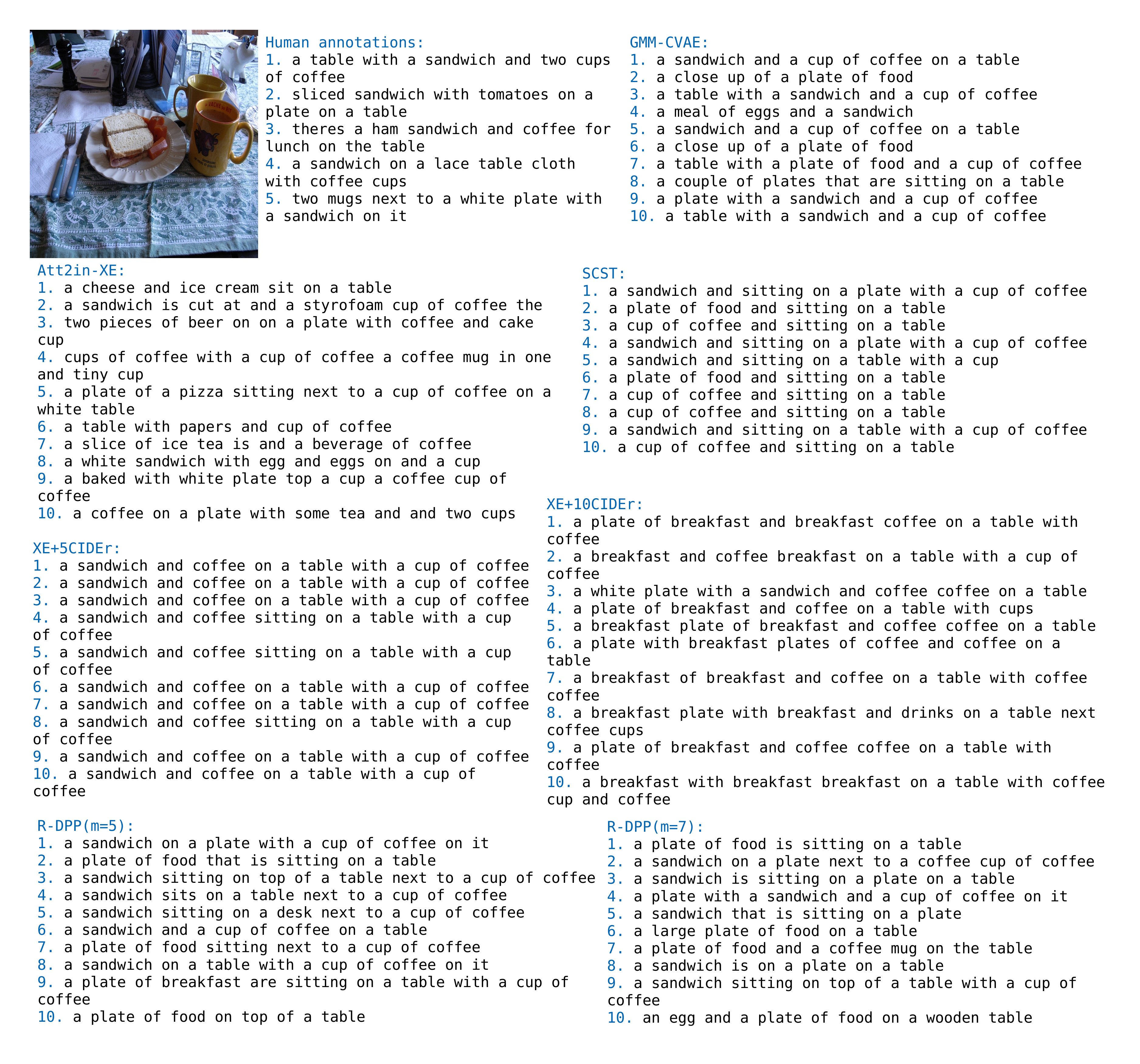}
\caption{•}\label{supp:fig3}
\end{figure*}

\begin{figure*}[t]
\centering
\includegraphics[width=\textwidth]{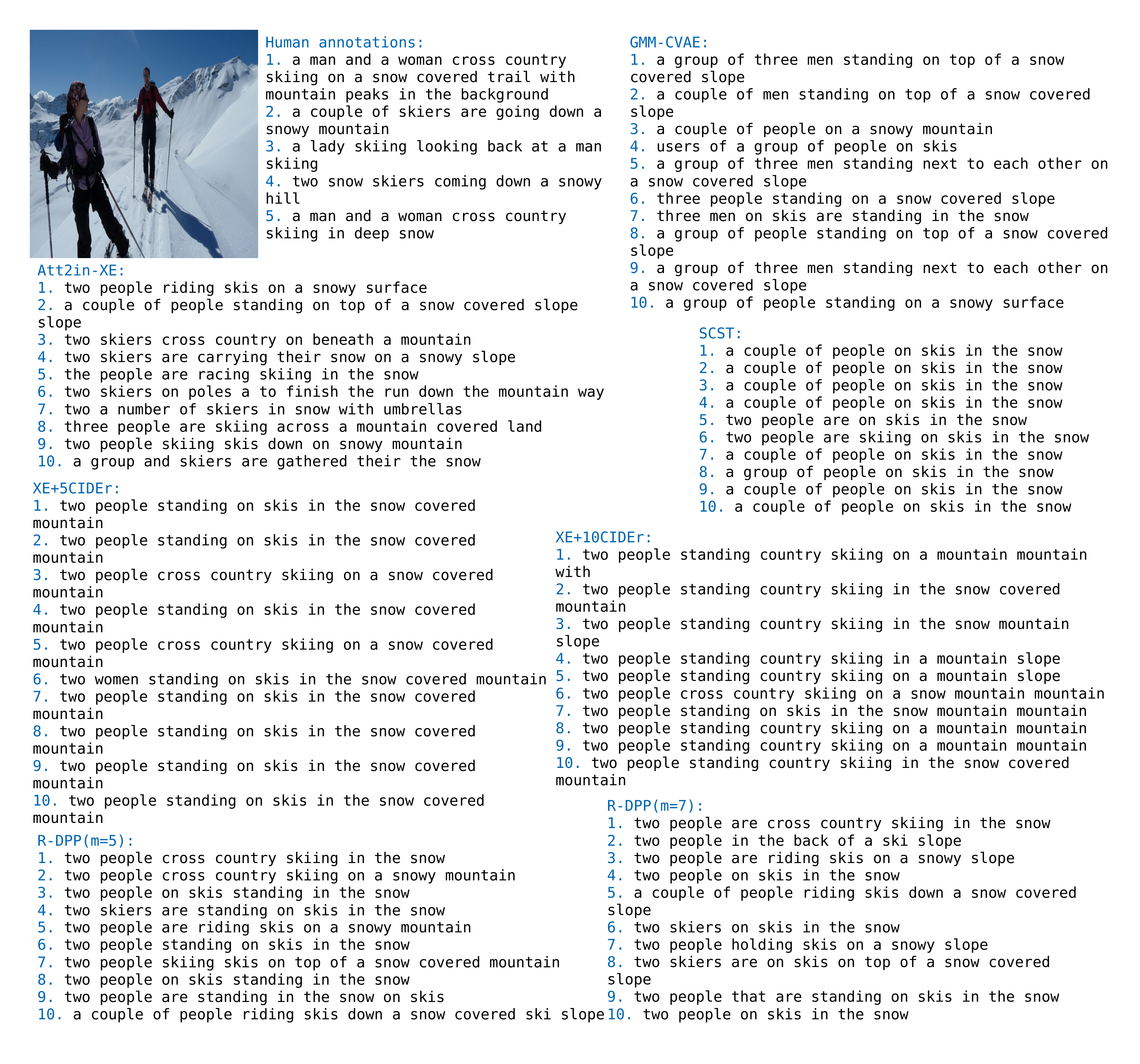}
\caption{•}\label{supp:fig4}
\end{figure*}

\begin{figure*}[t]
\centering
\includegraphics[width=\textwidth]{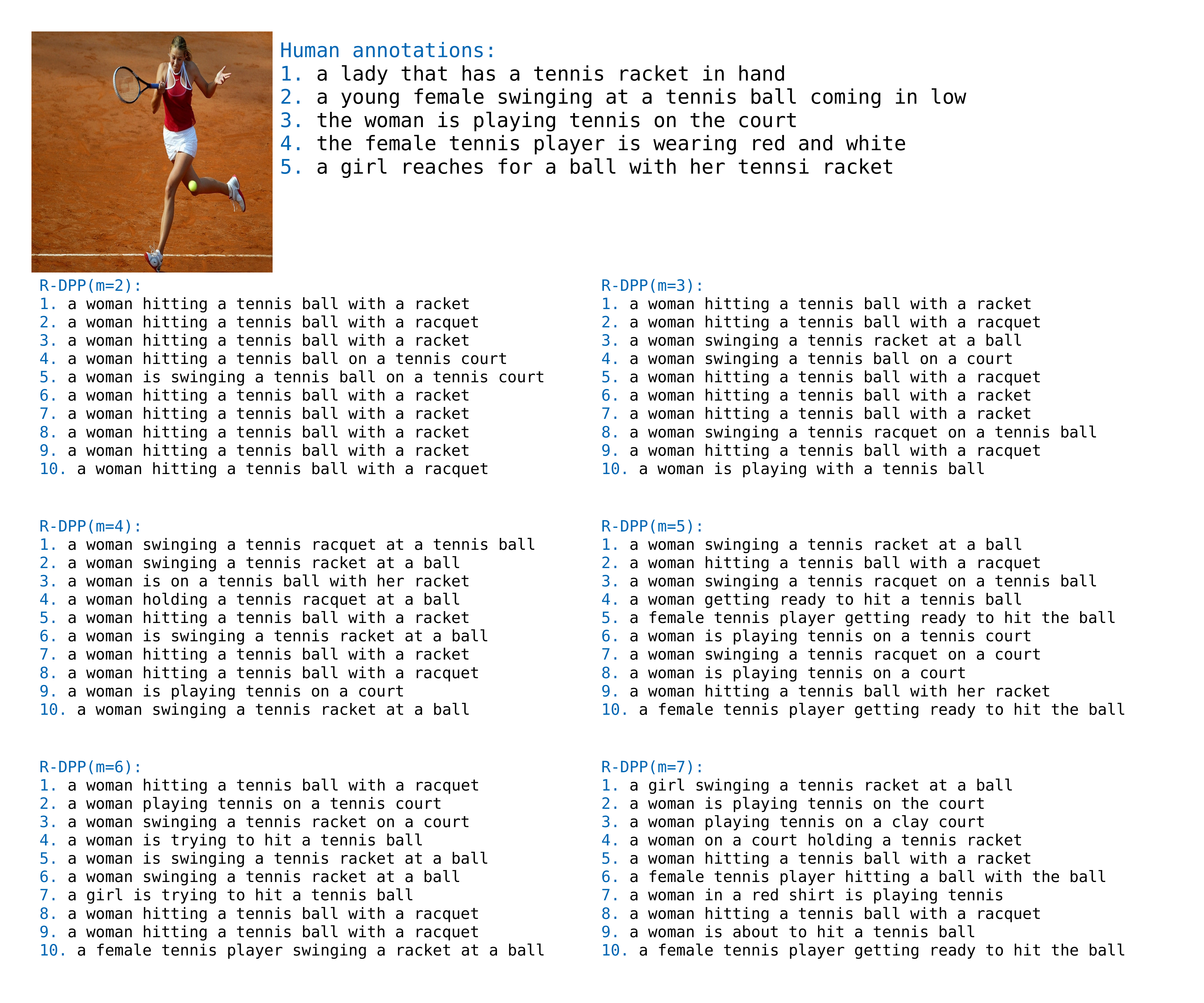}
\caption{•}\label{supp:fig5}
\end{figure*}

\begin{figure*}[t]
\centering
\includegraphics[width=\textwidth]{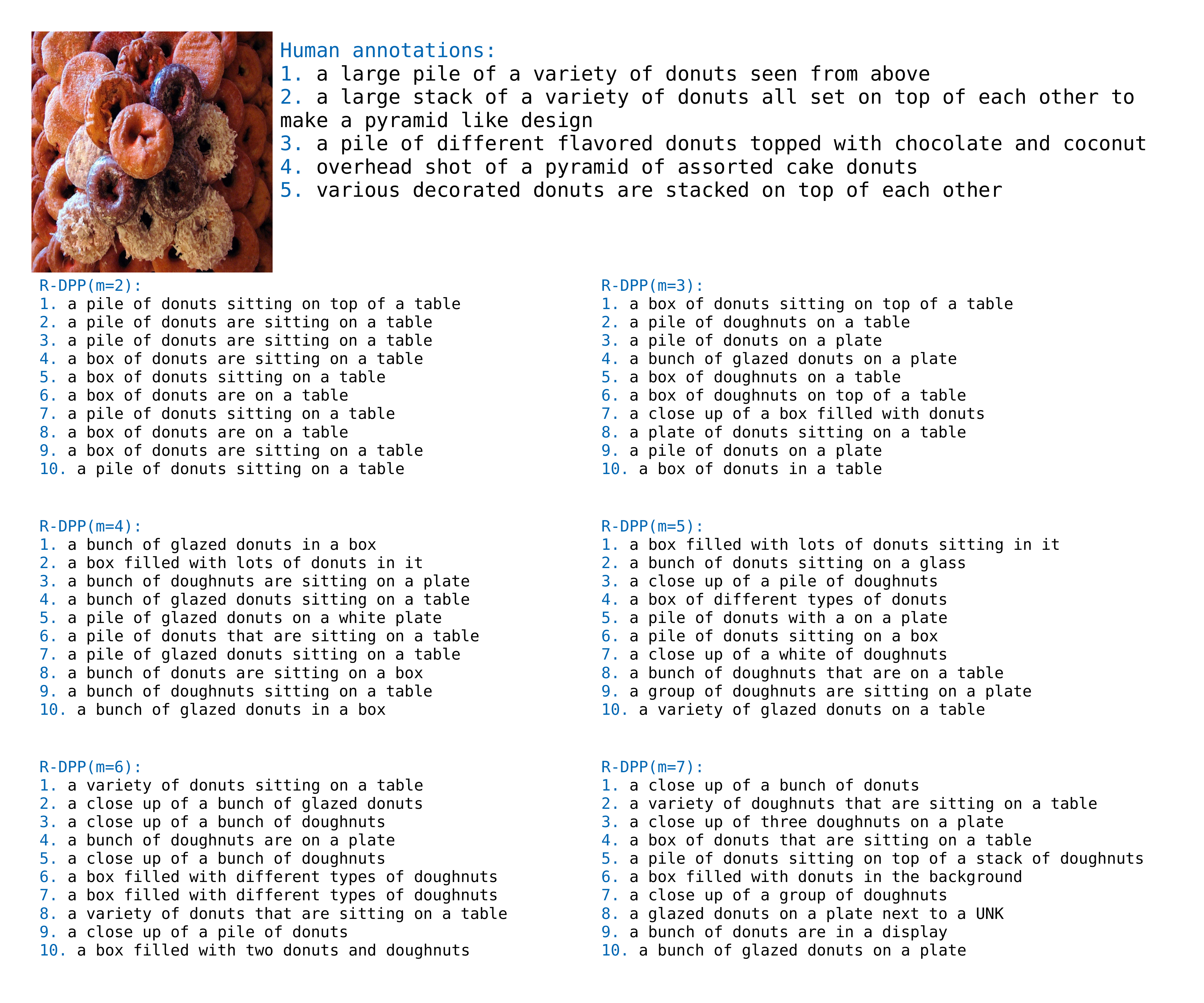}
\caption{•}\label{supp:fig6}
\end{figure*}

\begin{figure*}[t]
\centering
\includegraphics[width=\textwidth]{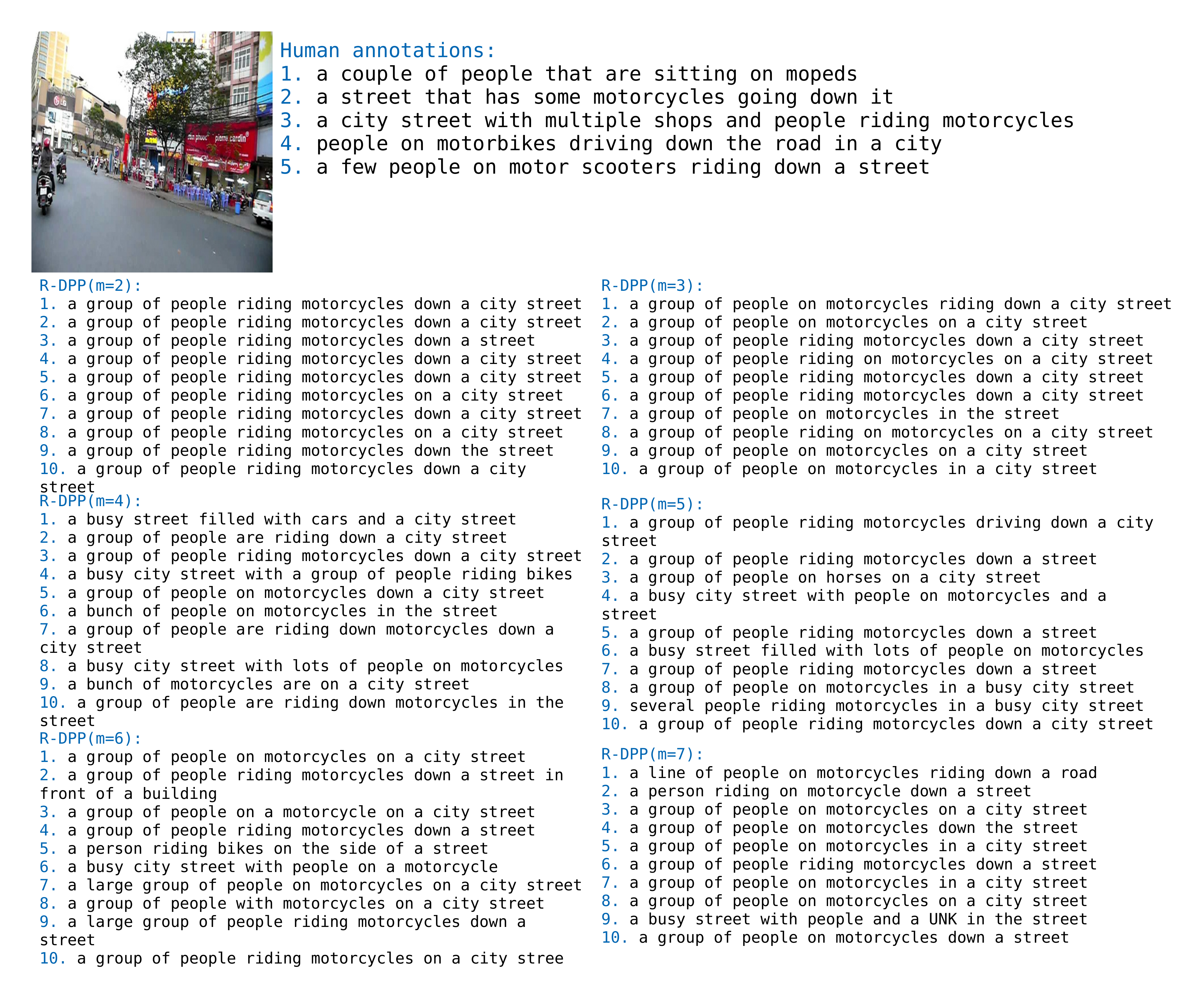}
\caption{•}\label{supp:fig7}
\end{figure*}

\begin{figure*}[t]
\centering
\includegraphics[width=\textwidth]{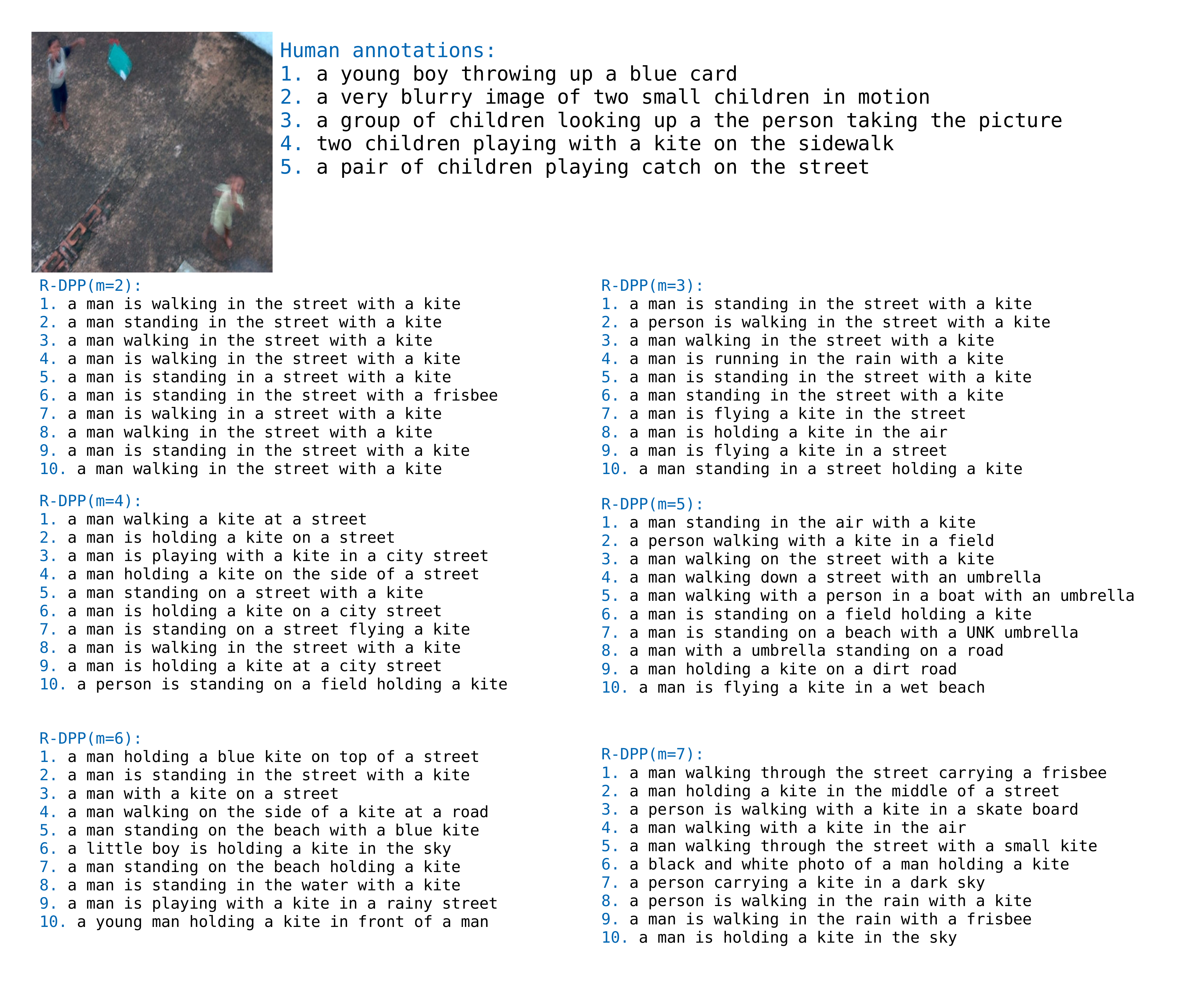}
\caption{•}\label{supp:fig8}
\end{figure*}

We show more qualitative results of R-DPP. Fig. \ref{supp:fig1} to \ref{supp:fig4} show the comparison between R-DPP and other models, and Fig. \ref{supp:fig5} to \ref{supp:fig8} show the generated captions by R-DPP with different numbers of samples during training. Compared with other methods, our R-DPP could generate more fluent and diverse captions. We find that R-DPP is able to generate captions with different sentence structures (syntactic diversity), such as using synonyms, redundant and concise descriptions.

\end{document}